\documentclass{article}

\usepackage[preprint,nonatbib]{neurips_2022}

\usepackage[utf8]{inputenc} 
\usepackage[T1]{fontenc}    
\usepackage{hyperref}       
\usepackage{url}            
\usepackage{booktabs}       
\usepackage{amsfonts}       
\usepackage{nicefrac}       
\usepackage{microtype}      
\usepackage{xcolor}         

\usepackage{wrapfig}
\usepackage{times}
\usepackage{latexsym}

\definecolor{darkblue}{rgb}{0.0,0.0,0.5}

\usepackage{amsmath}
\usepackage{array}
\newcolumntype{L}{>{$}l<{$}}
\newcolumntype{C}{>{$}c<{$}}
\newcolumntype{R}{>{$}r<{$}}

\usepackage{multirow}
\usepackage{tabularx, caption, boldline}
\usepackage{graphicx}
\usepackage{cellspace}
\usepackage{algpseudocode}  
\usepackage{multicol}  
\usepackage{flexisym}

\usepackage{microtype}

\usepackage{dsfont}

\makeatletter
\def\hlinewd#1{%
\noalign{\ifnum0=`}\fi\hrule \@height #1 %
\futurelet\reserved@a\@xhline}
\makeatother

\usepackage{times}
\usepackage{latexsym}

\usepackage{algpseudocode}  
\usepackage{amsmath}
\usepackage{multicol}  
\usepackage{multirow} 
\usepackage{graphicx}  
\usepackage{array}
\usepackage{arydshln}
\usepackage{booktabs}
\usepackage{textcomp}

\usepackage{amssymb}
\usepackage{pifont}

\usepackage{cleveref}
\crefformat{section}{\S#2#1#3} 
\crefformat{subsection}{\S#2#1#3}
\crefformat{subsubsection}{\S#2#1#3}

\usepackage{microtype}

\usepackage{adjustbox}
\usepackage{multicol}  
\usepackage{multirow} 
\usepackage{amsmath}
\usepackage{amsfonts}
\usepackage{makecell}
\usepackage{algpseudocode} 
\usepackage{verbatim}
\usepackage{mathtools}
\DeclareMathOperator*{\argmax}{arg\,max}

\usepackage{booktabs}

\usepackage{makecell}
\usepackage{cleveref}
\usepackage[normalem]{ulem}

\usepackage{listings}

\definecolor{Code}{rgb}{0,0,0} 
\definecolor{Decorators}{rgb}{0.5,0.5,0.5} 
\definecolor{Numbers}{rgb}{0.5,0,0} 
\definecolor{MatchingBrackets}{rgb}{0.25,0.5,0.5} 
\definecolor{Keywords}{rgb}{0,0,1} 
\definecolor{self}{rgb}{0,0,0} 
\definecolor{Strings}{rgb}{0,0.63,0} 
\definecolor{Comments}{rgb}{0.63,0,0} 
\definecolor{Backquotes}{rgb}{0,0,0} 
\definecolor{Classname}{rgb}{0,0,0} 
\definecolor{FunctionName}{rgb}{0,0,0} 
\definecolor{Operators}{rgb}{0,0,0} 
\definecolor{Background}{rgb}{0.99,0.99,0.99} 

\usepackage{setspace}
\usepackage{CJKutf8}

\usepackage{times}
\usepackage{latexsym}

\usepackage{footnote}
\makesavenoteenv{tabular}
\makesavenoteenv{table}

\usepackage[hang,flushmargin]{footmisc}

\usepackage[linesnumbered,ruled]{algorithm2e}
\SetAlFnt{\small}
\SetAlCapNameFnt{\normalsize}
\makeatletter
\newcommand{\nosemic}{\renewcommand{\@endalgocfline}{\relax}}
\newcommand{\dosemic}{\renewcommand{\@endalgocfline}{\algocf@endline}}
\let\oldnl\nl
\newcommand{\nonl}{\renewcommand{\nl}{\let\nl\oldnl}}
\makeatother

\usepackage{enumitem}
\usepackage{tablefootnote}

\usepackage{tabularx}
\makeatletter
\def\hlinewd#1{%
\noalign{\ifnum0=`}\fi\hrule \@height #1 %
\futurelet\reserved@a\@xhline}
\makeatother

\usepackage{minitoc}

\usepackage{cleveref}
\crefformat{section}{\S#2#1#3}
\crefformat{subsection}{\S#2#1#3}
\crefformat{subsubsection}{\S#2#1#3}
\crefrangeformat{section}{\S\S#3#1#4 to~#5#2#6}
\crefmultiformat{section}{\S\S#2#1#3}{ and~#2#1#3}{, #2#1#3}{ and~#2#1#3}

\usepackage{refstyle}
\Crefformat{figure}{#2Figure~#1#3}
\Crefmultiformat{figure}{Figures.~#2#1#3}{ and~#2#1#3}{, #2#1#3}{ and~#2#1#3}
\Crefformat{table}{#2Table~#1#3}
\Crefmultiformat{table}{Tables~#2#1#3}{ and~#2#1#3}{, #2#1#3}{ and~#2#1#3}
\Crefformat{appendix}{Appx.~\S#2#1#3}
\crefformat{algorithm}{Alg.~#2#1#3}

\definecolor{NavyBlue}{rgb}{0.1, 0.4, 0.8}
\hypersetup{colorlinks = true, linkcolor = NavyBlue,
            urlcolor  = gray,
            citecolor = NavyBlue,
            anchorcolor = NavyBlue}

\title{An Empirical Study On\\ Contrastive Search And Contrastive Decoding\\ For Open-ended Text Generation}

\author{%
  Yixuan Su$^\spadesuit$ \quad \quad \quad \quad \quad \quad \quad \quad Jialu Xu$^\heartsuit$\\
  $^\spadesuit$University of Cambridge\quad  \quad$^\heartsuit$Independent Researcher\\
  \texttt{ys484@cam.ac.uk} \\
}

\begin{document}
\maketitle


\begin{abstract}
In the study, we empirically compare the two recently proposed decoding methods, i.e. Contrastive Search (CS) and Contrastive Decoding (CD), for open-ended text generation. The automatic evaluation results suggest that, while CS performs worse than CD on the MAUVE metric, it substantially surpasses CD on the diversity and coherence metrics. More notably, extensive human evaluations across three different domains demonstrate that human annotators are universally more in favor of CS over CD with substantial margins. 

The contradicted results between MAUVE and human evaluations reveal that MAUVE does not accurately reflect human preferences. Therefore, we call upon the research community to develop better evaluation metrics for open-ended text generation. To ensure the reproducibility of our work, we have open-sourced all our code, evaluation results, as well as human annotations at \url{https://github.com/yxuansu/Contrastive_Search_versus_Contrastive_Decoding}.
\end{abstract}

\section{Introduction}
Open-ended text generation aims at generating coherent as well as informative text continuation based on the given prompt, and it is the core component in various NLP applications~\cite{su2022language,su2021prototype}. In this study, we compare the two recently proposed decoding methods for open-ended text generation, i.e. (i) contrastive decoding (CD)~\cite{li2022contrastive} and (ii) contrastive search (CS)~\cite{su2022a,su2022contrastive}.

For a comprehensive comparison, we follow Li \emph{et al.}~\cite{li2022contrastive} and conduct experiments on three benchmarks across different domains. On the one hand, the automatic evaluations (\cref{sec:automatic_evaluation}) indicate that CD performs notably better than CS on the MAUVE metric. However, CS achieves substantially better results on the diversity and coherence metrics. On the other hand, extensive human evaluations (\cref{sec:human_evaluation}) on three  benchmarks  validate that the human annotators are universally more in favor of the texts produced by CS than the ones produced by CD with substantial margins. 

Given the contradicted results of MAUVE and human evaluations, we argue that MAUVE does not accurately reflect human preferences. In \cref{sec:analysis}, we show that the human preferences better correlate with the balance between the diversity and the coherence aspects of the generated texts. Thereby, we suggest future research on better evaluation metrics for open-ended text generation to take into account these two aspects.



In summary, our contributions are:
\begin{itemize}
    \item We conduct comprehensive experiments to compare the two recently proposed decoding methods, i.e. CD and CS, for open-ended text generation.
    \item We demonstrate that MAUVE does not accurately reflect the human preferences on different methods for open-ended text generation. Moreover, we suggest a plausible direction for future research on better evaluation metrics of open-ended text generation.
\end{itemize}

\section{Preliminaries}
\subsection{Contrastive Decoding}

Contrastive decoding (CD) is introduced by Li \emph{et al.}~\cite{li2022contrastive}. Given a prompt text $\boldsymbol{x}_{<t}$, the selection of the output token $x_t$ is decided by comparing two separate language models (LM) as
\begin{equation}
    \label{eq:cd}
    x_t = \argmax_{v\in \mathcal{V}_{\textup{head}}(\boldsymbol{x}_{<t})}\bigg\{\log p_{\textup{EXP}}(v|\boldsymbol{x}_{<t})-\log p_{\textup{AMA}}(v|\boldsymbol{x}_{<t}, \tau)\bigg\},
\end{equation}
where $p_{\textup{EXP}}(\cdot|\boldsymbol{x}_{<t})$ is the probability distribution produced by an expert LM. The $p_{\textup{AMA}}(\cdot|\boldsymbol{x}_{<t}, \tau)$ is the probability distribution produced by an amateur LM scaled with a predefined temperature $\tau$. Typically, the expert LM (e.g. GPT2-XL) is larger than the amateur LM (e.g. GPT2-Small). The candidate set $\mathcal{V}_{\textup{head}}(\boldsymbol{x}_{<t})$ is defined as 
\begin{equation}
    \label{eq:cd_candidate}
    \mathcal{V}_{\textup{head}}(\boldsymbol{x}_{<t}) = \{v\in\mathcal{V}:p_{\textup{EXP}}(v|\boldsymbol{x}_{<t})\geq\alpha\times\max_{w}p_{\textup{EXP}}(w|\boldsymbol{x}_{<t})\},
\end{equation}
where $\alpha$ is a hyperparameter.

\subsection{Contrastive Search}
In contrast to CD, contrastive search (CS)~\cite{su2022a,su2022contrastive} only requires a single LM to generate the text continuation conditioned on the prompt. Formally, given the prompt text $\boldsymbol{x}_{<t}$, the selection of the output token $x_t$ follows
\begin{equation}
    \label{eq:score}
    x_t = \argmax_{v\in V^{(k)}}\bigg\{(1 - \alpha)\times \underbrace{p_{\theta}(v|\boldsymbol{x}_{<t})}_{\textup{model confidence}} -  \: \alpha \times \underbrace{(\max\{s(h_v, h_{x_j}):1\leq j \leq t-1\})}_{\textup{degeneration penalty}}\bigg\},
\end{equation}
where $V^{(k)}$ is the set of top-$k$ predictions from the LM's probability distribution $p_{\theta}(\cdot|\boldsymbol{x}_{<t})$. In Eq. (\ref{eq:score}), the first term, \textit{model confidence}, is the probability of the candidate $v$ predicted by the LM. The second term, \textit{degeneration penalty}, measures how discriminative of the candidate $v$ with respect to the previous context $\boldsymbol{x}_{<t}$ and $s(\cdot, \cdot)$ computes the cosine similarity between token representations. More specifically, degeneration penalty is defined as the maximum cosine similarity between the representation of the candidate $v$ and that of all tokens in $\boldsymbol{x}_{<t}$. Here, the candidate representation $h_v$ is computed by the LM given the concatenation of $\boldsymbol{x}_{<t}$ and $v$. Intuitively, a larger degeneration penalty of $v$ means it is more similar to the context, therefore more likely leading to the undesirable repetitions in the generated output. The hyperparameter $\alpha\in[0,1]$ regulates the importance of these two components.

\section{Experiment}

\textbf{Evaluation Benchmarks.} Following Li \emph{et al.}~\cite{li2022contrastive}, we conduct experiments on three benchmarks from different domains, including (i) articles from Wikinews\footnote{\url{http://www.wikinews.org/}} in the news domain; (ii) Wikitext-103 dataset~\cite{merity2016pointer} from the Wikipedia domain; (iii) and BookCorpus~\cite{zhu2015aligning} from the story domain. 

Same as in Li \emph{et al.}~\cite{li2022contrastive}, the generation of LM is conditioned on the test prompts with a fixed length of 32. And the generation of the text ends upon reaching an end-of-document token or a maximum length of 256 tokens. To ensure our experiments are aligned with Li \emph{et al.}~\cite{li2022contrastive}, we directly use the data provided in the authors' released repository\footnote{\url{https://github.com/XiangLi1999/ContrastiveDecoding/tree/main/text-generation/outputs_ignorePrefix_ccnews_256}}.

\textbf{Model and Baselines.} We compare different decoding methods using the GPT2-XL model~\cite{radford2019language}. (i) Following Li \emph{et al.}~\cite{li2022contrastive}, in contrastive decoding (CD), the expert and amateur LM are set as GPT2-XL and GPT2-Small, respectively; And the $\alpha$ (see Eq. (\ref{eq:cd_candidate})) and $\tau$ (see Eq. (\ref{eq:cd})) for CD  are set as $0.1$ and $0.5$, respectively. (ii) For contrastive search (CS), we set $\alpha$ (see Eq. (\ref{eq:score})) as a constant $0.6$; And $k$ (see Eq. (\ref{eq:score})) for news, Wikipedia, and story benchmarks is set as $5$, $5$, and $6$, respectively. 

In addition to CD and CS, in the experiments, we also report the results of other baseline methods, including (i) greedy search; (ii) top-$k$ sampling ($k=50$)~\cite{fan2018hierarchical}; (iii) nucleus sampling ($p=0.95$)~\cite{holtzman2019curious}; and (iv) typical sampling ($\tau=0.95$)~\cite{meister2022typical}. 

Note that, for a fair comparison with Li \emph{et al.}~\cite{li2022contrastive}, we report the performance of the baseline methods (i.e. greedy search, top-$k$ sampling, nucleus sampling, typical sampling, and contrastive decoding (CD)) using the generated texts provided in the authors' released repository\footnote{\url{https://github.com/XiangLi1999/ContrastiveDecoding/tree/main/text-generation/outputs_ignorePrefix_ccnews_256}}. However, for contrastive search (CS), the reported numbers in Li \emph{et al.}~\cite{li2022contrastive} are different from our reproduced numbers. Therefore, we re-implement the results of CS using the same benchmark data provided by Li \emph{et al.}~\cite{li2022contrastive} in their official repository\footnote{\url{https://github.com/XiangLi1999/ContrastiveDecoding/tree/main/text-generation/outputs_ignorePrefix_ccnews_256}}.

\subsection{Automatic Evaluation}
\label{sec:automatic_evaluation}

Following previous studies~\cite{su2022a,su2022contrastive,li2022contrastive}, we use the following metrics for automatic evaluation.

(i)\textbf{ Diversity} takes into account the generation repetition at different $n$-gram levels and it is defined as: $\textup{{diversity}}=\prod_{n=2}^{4}(1.0-\frac{\textup{rep-n}}{100})$, where $\textup{{rep-n}}=100 \times (1.0 - \frac{|\textup{unique n-grams}(\hat{\boldsymbol{x}})|}{|\textup{total n-grams}(\hat{\boldsymbol{x}})|})$ and $\hat{\boldsymbol{x}}$ is the text generated by the LM. 

(ii)\textbf{ MAUVE}~\cite{pillutla2021mauve} is designed for measuring the token distribution closeness between the generated text and the human-written text over the whole test set. Note that, while the maximum length of generation in the experiments is 256, we follow Li \emph{et al.}~\cite{li2022contrastive} and measure the MAUVE score by truncating the generated text to its first 128 tokens.

(iii)\textbf{ Coherence} is recently introduced by Su and Collier~\cite{su2022contrastive} and it automatically measures the semantic coherence between the prompt and the generated text. Formally, given the prompt $\boldsymbol{x}$ and the generated text $\hat{\boldsymbol{x}}$, coherence is defined as the averaged log-likelihood of $\hat{\boldsymbol{x}}$ conditioned on $\boldsymbol{x}$ as
\begin{equation}
    \label{eq:coherence}
    \textup{coherence}(\hat{\boldsymbol{x}}, \boldsymbol{x}) = \frac{1}{|\hat{\boldsymbol{x}}|}\sum_{i=1}^{|\hat{\boldsymbol{x}}|}\log p_{\mathcal{M}}(\hat{\boldsymbol{x}}_{i}|[\boldsymbol{x}:\hat{\boldsymbol{x}}_{<i}]),
\end{equation}
where  [:] is the concatenation operation and $\mathcal{M}$ is a massively pre-trained LM. In our experiments, we follow Su and Collier~\cite{su2022contrastive} and set $\mathcal{M}$ as the OPT-2.7B model~\cite{zhang2022opt}.

\begin{table*}[h]
    \small
	\centering  
	\renewcommand{\arraystretch}{1.2}
	\setlength{\tabcolsep}{6pt}
	\scalebox{0.83}{
	\begin{tabular}{cccccccccc}
	    \hlinewd{0.75pt}
	    \multirow{2}{*}{\textbf{Method}}&\multicolumn{3}{c}{\textbf{Wikinews}}&\multicolumn{3}{c}{\textbf{Wikitext}}&\multicolumn{3}{c}{\textbf{Story}}\\
	    \cmidrule(lr){2-4}
	    \cmidrule(lr){5-7}
	    \cmidrule(lr){8-10}
	    &div.(\%)&MAUVE(\%)&coh.&div.(\%)&MAUVE(\%)&coh.&div.(\%)&MAUVE(\%)&coh.\\
	    \hline
	    Greedy Search$^{*}$&3.55&13.96&\textbf{-0.47}&1.77&4.91&\textbf{-0.41}&0.86&2.65&\textbf{-0.34}\\
	    Top-$k$ Sampling$^{*}$&91.56&89.86&-2.22&87.49&81.00&-2.37&91.22&87.49&-2.45\\
	    Nucleus Sampling$^{*}$&93.54&89.45&-2.61&92.16&86.54&-3.03&94.50&91.47&-3.02\\
	    Typical Sampling$^{*}$&\textbf{95.37}&90.97&-3.26&\textbf{94.82}&86.07&-3.71&\textbf{96.29}&88.58&-3.68\\
	    Contrastive Decoding$^{*}$&91.57&\textbf{92.20}&-2.16&88.02&\textbf{91.46}&-2.19&86.41&\textbf{93.17}&-2.09\\
	    Contrastive Search&93.72&84.14&-1.39&89.35&77.97&-1.56&93.06&84.74&-1.61\\
		\hlinewd{0.75pt}
	\end{tabular}}
    \caption{Automatic evaluation results, where div. and coh. denote diversity and coherence. The numbers marked with $^{*}$ are obtained using the generated texts originally released by Li \emph{et al.}~\cite{li2022contrastive}.}
    	\vspace{-1.5mm}
	\label{tb:main_results}
\end{table*}

\textbf{Evaluation Results.} Table~\ref{tb:main_results} presents the automatic evaluation results. On the one hand, we see that CD achieves the best MAUVE score on all evaluated benchmarks. On the other hand, CS yields competitive performances on the diversity metric and achieves substantially better results on the coherence metric than CD and other sampling methods.

\subsection{Human Evaluation}
\label{sec:human_evaluation}
To further compare contrastive decoding (CD) with contrastive search (CS), we conduct a human evaluation with 4 native-speaker graders from a third-party grading platform. We randomly select 150 test prompts from the benchmarks across different domains, and evaluate CD and CS through pairwise comparison. Specifically, for each test prompt, the annotators are given two texts, with random order, that are generated by CD and CS. The annotators then decide which one is more likely written by humans considering the following aspects of the generated text:

\begin{itemize}
    \item \textbf{Coherence}: Whether the generated text is semantically coherent.
    \item \textbf{Fluency}: Whether the generated text is fluent and easy to understand.
    \item \textbf{Informativeness}: Whether the generated text is diverse and contains interesting content. 
\end{itemize}

\begin{table*}[h]
    \small
	\centering  
	\renewcommand{\arraystretch}{1.2}
	\setlength{\tabcolsep}{6pt}
	\scalebox{1.0}{
	\begin{tabular}{cccccc}
	    \hlinewd{0.75pt}
	    \multirow{5}{*}{\rotatebox[origin=c]{90}{{\textbf{Wikinews}}}}&
	    \multicolumn{2}{c}{Method A is better}&Neutral&\multicolumn{2}{c}{Method B is better}\\
	    \cmidrule(lr){2-3}
	    \cmidrule(lr){4-4}
	    \cmidrule(lr){5-6}
	    &Nucleus Sampling$^{*}$&25.0\%&4.2\%&\textbf{70.8}\%$^{\dagger}$&Contrastive Decoding$^{*}$\\
	    &Typical Sampling$^{*}$&7.8\%&15.1\%&\textbf{77.1}\%$^{\dagger}$&Contrastive Decoding$^{*}$\\
	    &Contrastive Search&\textbf{68.5}\%$^{\dagger}$&2.0\%&29.5\%&Contrastive Decoding\\
	    \hlinewd{0.75pt}
	    \multirow{5}{*}{\rotatebox[origin=c]{90}{{\textbf{Wikitext}}}}&
	    \multicolumn{2}{c}{Method A is better}&Neutral&\multicolumn{2}{c}{Method B is better}\\
	    \cmidrule(lr){2-3}
	    \cmidrule(lr){4-4}
	    \cmidrule(lr){5-6}
	    &Nucleus Sampling$^{*}$&20.2\%&8.3\%&\textbf{71.4}\%$^{\dagger}$&Contrastive Decoding$^{*}$\\
	    &Typical Sampling$^{*}$&6.7\%&4.6\%&\textbf{88.7}\%$^{\dagger}$&Contrastive Decoding$^{*}$\\
	    &Contrastive Search&\textbf{65.0}\%$^{\dagger}$&2.0\%&33.0\%&Contrastive Decoding\\
	    \hlinewd{0.75pt}
	    \multirow{5}{*}{\rotatebox[origin=c]{90}{{\textbf{Story}}}}&
	    \multicolumn{2}{c}{Method A is better}&Neutral&\multicolumn{2}{c}{Method B is better}\\
	    \cmidrule(lr){2-3}
	    \cmidrule(lr){4-4}
	    \cmidrule(lr){5-6}
	    &Nucleus Sampling$^{*}$&31.8\%&4.5\%&\textbf{63.6}\%$^{\dagger}$&Contrastive Decoding$^{*}$\\
	    &Typical Sampling$^{*}$&23.8\%&25.6\%&\textbf{50.6}\%$^{\dagger}$&Contrastive Decoding$^{*}$\\
	    &Contrastive Search&\textbf{67.0}\%$^{\dagger}$&1.0\%&32.0\%&Contrastive Decoding\\
		\hlinewd{0.75pt}
	\end{tabular}}
    \caption{Human evaluation results. $^{\dagger}$ means one method performs significantly better than the other as judged by Sign Test with $p$-value < 0.05. $^{*}$ the pairwise evaluation results between (i) nucleus sampling and contrastive decoding as well as (ii) typical sampling and contrastive decoding are directly cited from Li \emph{et al.}~\cite{li2022contrastive}.}
	\label{tb:human_evaluation}
\end{table*}

Table~\ref{tb:human_evaluation} presents the human evaluation results which validate that contrastive search (CS) significantly outperforms contrastive decoding (CD) and other sampling methods\footnote{In Li \emph{et al.}~\cite{li2022contrastive}, the authors compare contrastive decoding with nucleus and typical sampling. In Table~\ref{tb:human_evaluation}, we directly cite their results of human evaluation.} in all evaluated benchmarks from different domains. These results clearly demonstrate the superiority of contrastive search over other existing decoding strategies.

It is worth emphasizing that, as shown in Table~\ref{tb:main_results}, contrastive search yields notably lower MAUVE scores than CD and other sampling methods. Given this clear contradiction between MAUVE and human evaluations, we argue that MAUVE does not accurately reflect human preferences. Therefore, we call upon the research community to develop better evaluation metrics, for open-ended text generation, that more correlates with human judgements.

\subsection{Case Study}
Table~\ref{tb:case_examples} presents a qualitative example, from the news domain, comparing contrastive decoding and contrastive search. We see that the text generated by contrastive decoding contains excessive repetitions both on the lexical and phrasal levels, e.g. \textit{``The Pentagon”}, \textit{``The drones would likely”}, and etc. In contrast, the text generated by contrastive search is semantically coherent as well as grammatically fluent. It elaborates on the reasons of the military strike and provides diverse details of the incident. In Appendix~\ref{appendix:case_study}, we provide more qualitative examples for the comparison between these two decoding methods.

\begin{table*}[t]
    \small
	\centering  
	\renewcommand{\arraystretch}{1.2}
	\setlength{\tabcolsep}{6pt}
	\scalebox{0.82}{
	\begin{tabular}{cc}
	    \hlinewd{0.75pt}
        \textbf{Method}&\textbf{Generated Result}\\
        \hline
        Contrastive Decoding&\makecell[l]{\underline{\textbf{The first attacks carried out by United States Predator drones in Libya reportedly occurred}}\\ \underline{\textbf{today, as the Pentagon confirmed a strike carried out by the U.S. Air Force}} targeting "a senior\\ ISIL leader" near Benghazi. The strike reportedly killed Abu Yahia, an Egyptian who is the "emir of\\ external operations" for ISIL. \textcolor{red}{The Pentagon} also confirmed a second strike in Libya today targeting\\ Abu Anas al Libi. \textcolor{red}{The Pentagon} has yet to release a casualty count for either strike, though it's likely\\ that both killed a significant number. The strikes come after \textcolor{red}{the Pentagon} admitted to killing a Libyan\\ militant leader in a strike earlier today, though the militant leader was reportedly not the target of\\ today's drone attack. The drone strikes in Libya come as \textcolor{red}{the Pentagon} is reportedly \textcolor{red}{considering}\\ \textcolor{red}{sending} additional drones to Libya. \textcolor{red}{The Pentagon} is currently \textcolor{red}{considering sending} up to six Predator B\\ unmanned aircraft to Libya to conduct surveillance missions against Islamic militants there, according a\\ report from the Washington Times. \textcolor{red}{The drones would likely} operate out of the nearby Naval Air Station\\ Sigonella in Sicily. \textcolor{red}{The drones would} reportedly operate in the country for a year before being replaced\\ by manned aircraft. \textcolor{red}{The drones would likely} be used in a support capacity to manned aircraft, and not to\\ conduct strikes on their own, though \textcolor{red}{the drones would likely} be armed and \textcolor{red}{could potentially carry}\\ Hellfire \textcolor{red}{missiles}, which are currently in limited supply for use by manned aircraft in the country, though\\ \textcolor{red}{the drones could potentially carry} the \textcolor{red}{missiles}...}\\
        \hline
        Contrastive Search&\makecell[l]{\underline{\textbf{The first attacks carried out by United States Predator drones in Libya reportedly occurred}}\\ \underline{\textbf{today, as the Pentagon confirmed a strike carried out by the U.S. Air Force}} in Sirte.\\\\According to Reuters, the drone was targeting militants who "appeared to be preparing to attack a\\ vehicle carrying Westerners, possibly American, on the outskirts of the city." It is unclear what the\\ target was, but Reuters reports that "the attack was the first time the United States has acknowledged\\ using drones in Libya since the overthrow of Muammar Gaddafi in 2011."\\\\In an interview with The New York Times last week, Secretary of State John Kerry said, "We are going\\ to continue to take strikes wherever we can get them, but the fact of the matter is there are no boots on\\ the ground in Libya right now, which is a good thing for our national security and it's a good thing for\\ people's lives."\\\\The Pentagon has been reluctant to provide details about the strikes, citing operational security concerns.\\ However, a Pentagon spokesman, Rear Adm. John Kirby, said in a statement that "U.S. forces conducted\\ a strike against what they believed to be al-Qa'ida-associated terrorists in Sirte, Libya, on April 6,\\ targeting a location used by terrorists to train for attacks against Westerners."\\\\Kirby did not confirm the identity of the strike, but said it was...}\\
		\hlinewd{0.75pt}
	\end{tabular}}
    \caption{Texts generated by contrastive decoding and contrastive search given the same \underline{\textbf{prompt text}}. On the one hand, the text generated by contrastive decoding contains repetitions at the lexical and phrasal levels as highlighted in \textcolor{red}{red}. On the other hand, the text generated by contrastive search is semantically coherent as well as grammatically fluent. (Best viewed in color and by zooming in.)}
	\label{tb:case_examples}
\end{table*}

\section{Further Analysis}
\label{sec:analysis}

In this section, we provide in-depth comparison between contrastive search and other decoding methods. Specifically, we vary the $k$ (see Eq. (\ref{eq:score})), from 2 to 10, in contrastive search\footnote{The $\alpha$ (see Eq. (\ref{eq:score})) in contrastive search is kept as a constant 0.6.} to generate texts using the benchmark from the Wikipedia domain. The generated texts are evaluated from three aspects, i.e. (i) coherence; (ii) diversity; and (iii) MAUVE, which are described in  \cref{sec:automatic_evaluation}.

\begin{figure}[tb]
\centering
\begin{minipage}{.5\textwidth}
  \centering
  \includegraphics[width=1.\linewidth]{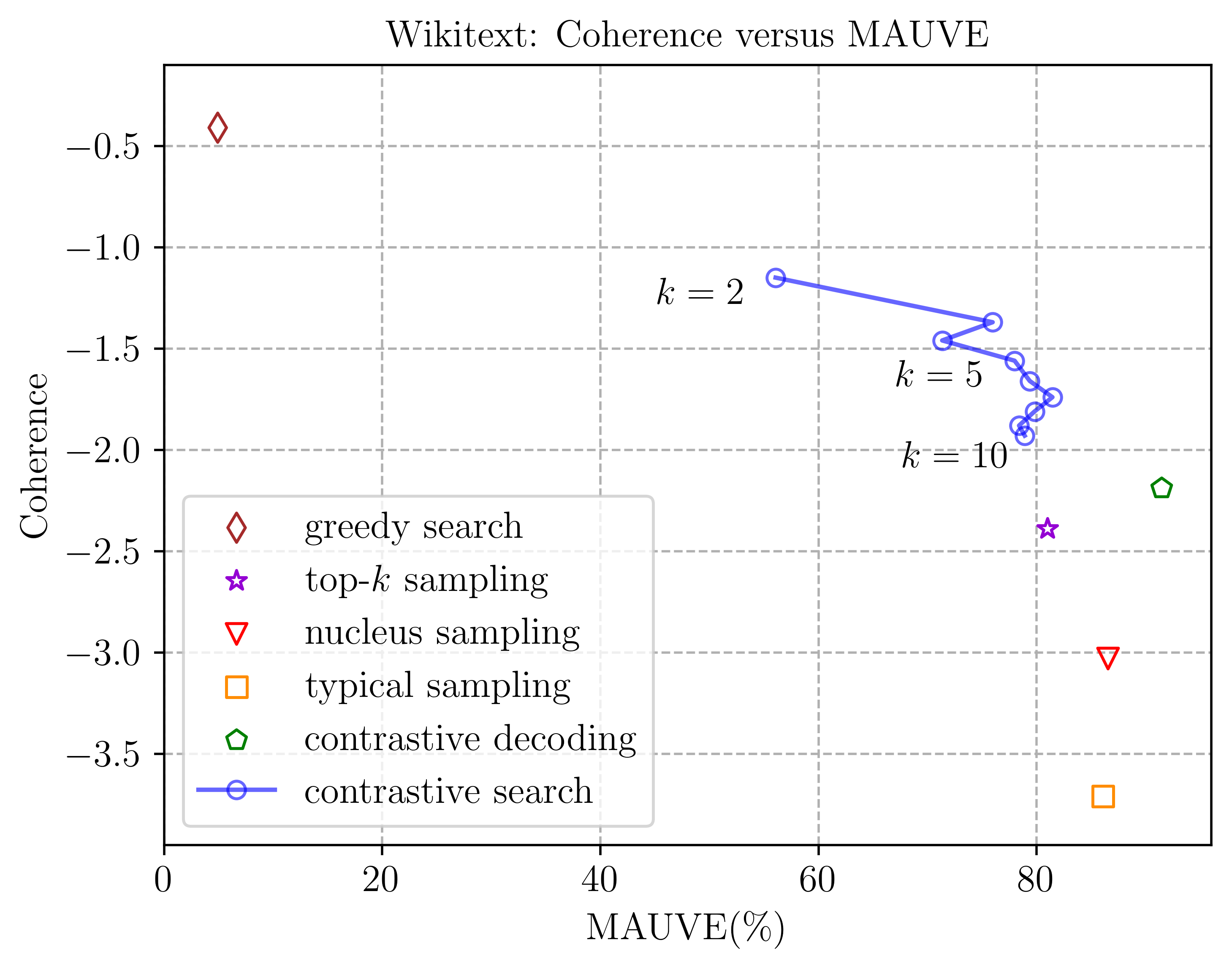}
  \captionof{figure}{\textbf{Wikitext} - Coherence versus MAUVE.}
  \label{fig:wikitext_coherence_vs_mauve}
\end{minipage}%
\begin{minipage}{.5\textwidth}
  \centering
  \includegraphics[width=1.\linewidth]{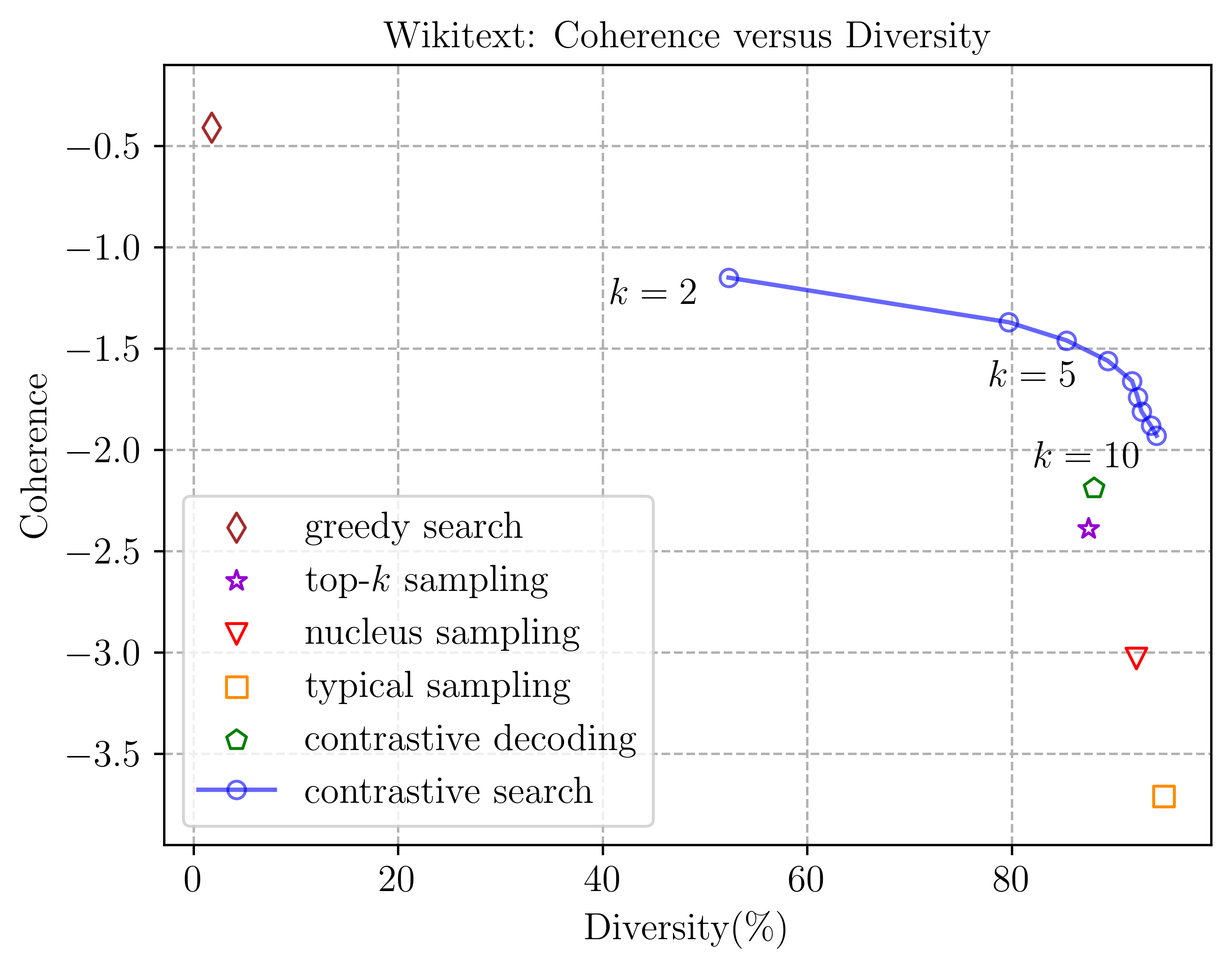}
  \captionof{figure}{\textbf{Wikitext} - Coherence versus Diversity.}
  \label{fig:wikitext_coherence_vs_diversity}
\end{minipage}
\end{figure}

The evaluated results are presented in Figure~\ref{fig:wikitext_coherence_vs_mauve} and Figure~\ref{fig:wikitext_coherence_vs_diversity}, respectively.\footnote{The results on benchmarks from the news and story domains can be found in Appendix~\ref{appendix:decoding_methods_comparison}.} On the one hand, Figure~\ref{fig:wikitext_coherence_vs_mauve} indicates that the MAUVE score of contrastive search lags behind other decoding methods (except for greedy search) with clear margins, which obviously contradicts to the human judgements as presented in~\cref{sec:human_evaluation}. Even by jointly considering the coherence and MAUVE metrics, it is hard to identify the better decoding method. On the other hand, from Figure~\ref{fig:wikitext_coherence_vs_diversity}, we see that contrastive search notably outperforms other methods on the balance between the coherence and diversity metrics, better correlating with human judgements. 

Our results demonstrate that MAUVE does not accurately reflect human preferences on different methods. Moreover, we suggest future research on better evaluation metrics, for open-ended text generation, to take into account both the coherence and the diversity aspects of the generated text.

\section{Conclusion}
In this work, we empirically compare the two recently proposed decoding methods, i.e. contrastive decoding (CD) and contrastive search (CS). We conduct extensive experiments on three benchmarks from different domains. The automatic evaluation results suggest that CD achieves better results on MAUVE while CS performs better on diversity and coherence. Moreover, through extensive human evaluations, we show that the human annotators are universally more in favor of CS over CD with substantial margins. Given the contradicted results between MAUVE and human evaluations, we provide in-depth analysis which reveals that the balance between the diversity and coherence metrics better correlates with human judgements. Our observation provides a plausible path for future research on better evaluation metrics for open-ended text generation.


\bibliographystyle{plain}
\bibliography{reference}

\appendix

\section{More Qualitative Examples}
\label{appendix:case_study}

In Table~\ref{tb:case_example_2} and Table~\ref{tb:case_example_3}, we provide more qualitative examples to compare contrastive decoding and contrastive search.

\begin{table*}[h]
    \small
	\centering  
	\renewcommand{\arraystretch}{1.2}
	\setlength{\tabcolsep}{6pt}
	\scalebox{0.8}{
	\begin{tabular}{cc}
	    \hlinewd{0.75pt}
        \textbf{Method}&\textbf{Generated Result}\\
        \hline
        Contrastive Decoding&\makecell[l]{\underline{\textbf{The first North Korean crossing at the Paekchin ferry caught the Heavy Mortar Platoon unprepared}}\\ \underline{\textbf{in the act of setting up its weapons. It also caugh}}t them completely unprepared to defend the crossing\\ point against a determined North Korean infantry attack, and it caught the entire platoon completely\\ unaware that a second crossing had been completed by another North Korean unit, the 2d North Korean\\ Tank Regiment (TKDR). This is not a story about a single event, or about one particular platoon or platoon\\ company, or even a particular company or company company company, platoon, or company company\\ platoon, platoon company, company platoon, company company platoon company platoon company\\ company, company company, company, or company company, company platoon company platoon company\\ platoon company, or even a particular platoon or company. This is about an entire North Korean division,\\ the 1st Tank Army (TKAD) and a divisional commander who was a tank officer, and the entire division and\\ his staff and his entire tank company and company company company and company platoon and company\\ company platoon company and company company company company and company company company\\ company company company company company. The story begins in late August, 1950. The division had\\ just crossed the 38 th parallel into South Korean-held territory from North Korean territory. This crossing\\ had been accomplished without the divisional artillery being fired, without any divisional infantry being\\ sent across, and with the divisional engineer unit not even arriving until September...}\\
        \hline
        Contrastive Search&\makecell[l]{\underline{\textbf{The first North Korean crossing at the Paekchin ferry caught the Heavy Mortar Platoon unprepared}}\\ \underline{\textbf{in the act of setting up its weapons. It also caugh}} to the fact that the North Koreans were able to get a\\ hold of an M1 Garand and a.30-06, two of the most commonly used weapons in the U.S. Army and\\ Marine Corps.\\\\This is the first time in my career that I have been in a situation where a weapon of this calibre was used in\\ combat. It was a big shock to all of us and we had no idea what to do with the weapons that we had in\\ our hands.\\\\The Heavy Mortar Platoon, under the command of Lt. Cmdr. John Bohn, took cover behind a sandbagged\\ bunker and waited for the attack to come. Bohn's men fired at the North Koreans as they approached the\\ bunkers, but their ammunition ran out before they could do much damage.\\\\After several minutes of fire, the bunker was breached and the platoon made a hasty retreat to the safety\\ of the bunker. As soon as they were out of harm's way, a mortar round slammed into the bunker and blew\\ it to smithereens.\\\\"That was one of the most intense moments of my life," Bohn said. "I remember looking at the smoke\\ billowing out of the hole and thinking...}\\
		\hlinewd{0.75pt}
	\end{tabular}}
    \caption{Qualitative example, from the Wikipedia domain, comparing contrastive decoding and contrastive search. (Best viewed by zooming in.)}
	\label{tb:case_example_2}
\end{table*}

\clearpage

\begin{table*}[t]
    \small
	\centering  
	\renewcommand{\arraystretch}{1.2}
	\setlength{\tabcolsep}{6pt}
	\scalebox{0.8}{
	\begin{tabular}{cc}
	    \hlinewd{0.75pt}
        \textbf{Method}&\textbf{Generated Result}\\
        \hline
        Contrastive Decoding&\makecell[l]{\underline{\textbf{Segun was completely lost. His system was on fire. How could he escape this mountain that had come }}\\ \underline{\textbf{to Mohamed even when Mohamed never gave it a thought to}} flee? He could only think to escape from\\ this mountain that had come to him even when Mohamed never gave it a thought to flee? He was completely\\ lost. His system was on fire.\\\\"Ahh…….I am lost…….I cannot even find the way…….Ahhh, I cannot escape……I cannot……I…\\…I am……lost……I can……I can only think……to……to……escape……from……from……\\from this……mountains……mountain that……that……mountain came…… to…… me…..me…..\\me……me…..me….."\\\\Ahhhhhhhhhhh……..I am……lost….. I…… I……I…… I……I…… I……I…… I…… I can\\ only…… to……escape….. from….. from….. this…..this……this……mountain…… mountain…..\\ that….. that….. mountain came…..to….. me….. me….. me….. me….. me……. I……I…… I…\\…I…… I……I…… I……I…… I……I…… I……I…… I……I…… I……I……\\\\"I cannot…….I……. I…….I……. I……. I……. I…… I cannot escape……. I...}\\
        \hline
        Contrastive Search&\makecell[l]{\underline{\textbf{Segun was completely lost. His system was on fire. How could he escape this mountain that had come }}\\ \underline{\textbf{to Mohamed even when Mohamed never gave it a thought to}}  do so?\\\\Mohamed looked at the mountain in front of him and thought, "This is the most important thing in my life\\ right now. I need to climb this mountain as soon as possible so no one gets the chance to catch me."\\\\As a result, he started the climb at 5:30 in the morning without sleeping for three days and three nights. It\\ was a grueling and tepid climb with little to no food and very little water. The temperature hovered around\\ 30 degrees centigrade (86 degrees Fahrenheit), and there was nothing but wind blowing all around him.\\\\After climbing the 1,000 meters (3,280 feet) in less than an hour, Mohamed reached the top of the mountain.\\ He felt like he had accomplished something, but at the same time, he was in a daze. He had no idea what to\\ do with his newly acquired power of telekinesis.\\\\Mohamed took a deep breath and said, "What the hell am I doing?"\\\\The rest of the day was spent trying to figure out how to get back down to the base camp and what was going\\ to happen to him if he made it back alive.\\\\At one point, a group of...}\\
		\hlinewd{0.75pt}
	\end{tabular}}
    \caption{Qualitative example, from the story domain, comparing contrastive decoding and contrastive search. (Best viewed by zooming in.)}
	\label{tb:case_example_3}
\end{table*}

\section{Comparison between Contrastive Search and other Decoding Methods}
\label{appendix:decoding_methods_comparison}
\begin{figure}[h]
\centering
\begin{minipage}{.5\textwidth}
  \centering
  \includegraphics[width=1.\linewidth]{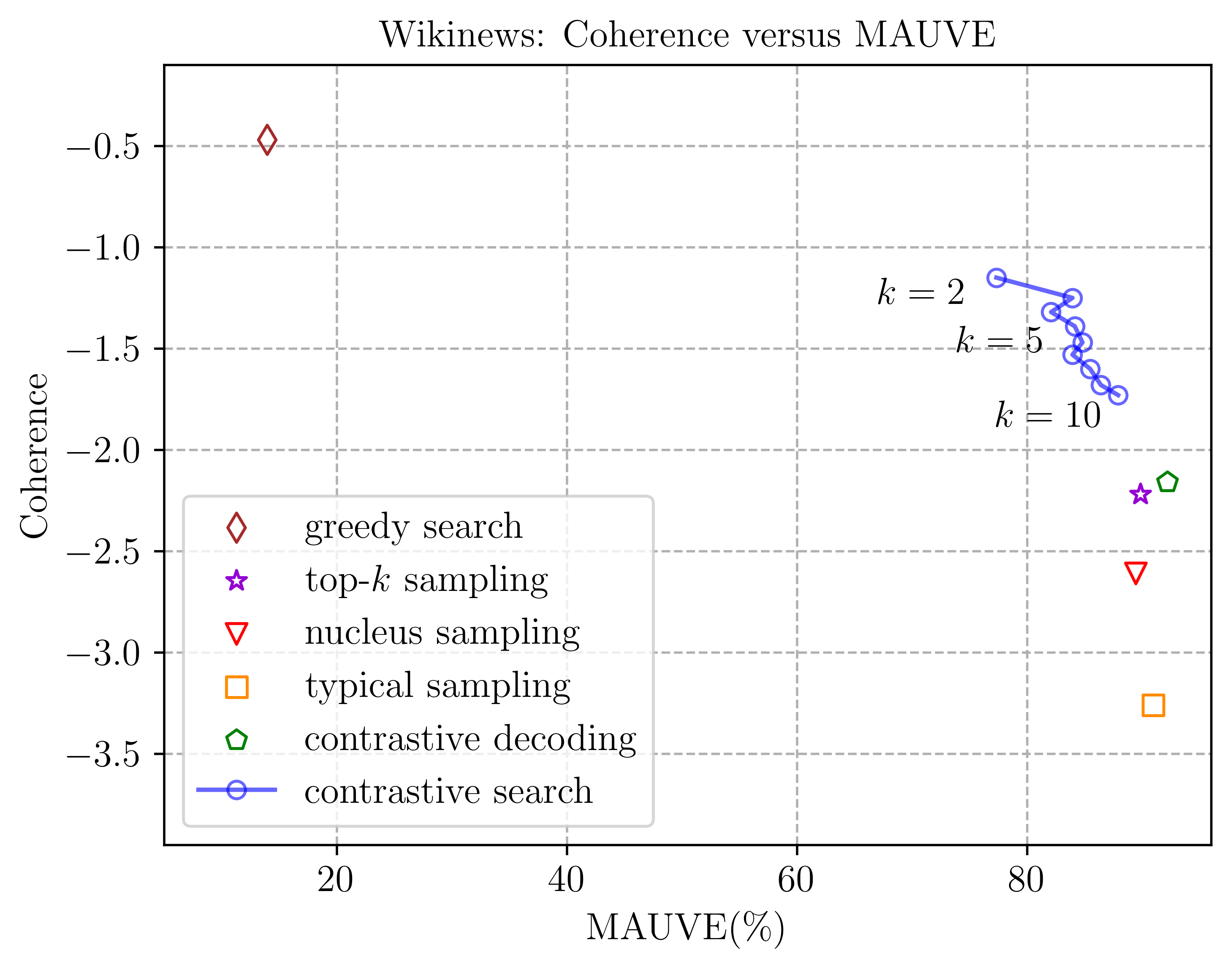}
  \captionof{figure}{\textbf{Wikinews} - Coherence versus MAUVE.}
  \label{fig:wikinews_coherence_vs_mauve}
\end{minipage}%
\begin{minipage}{.5\textwidth}
  \centering
  \includegraphics[width=1.\linewidth]{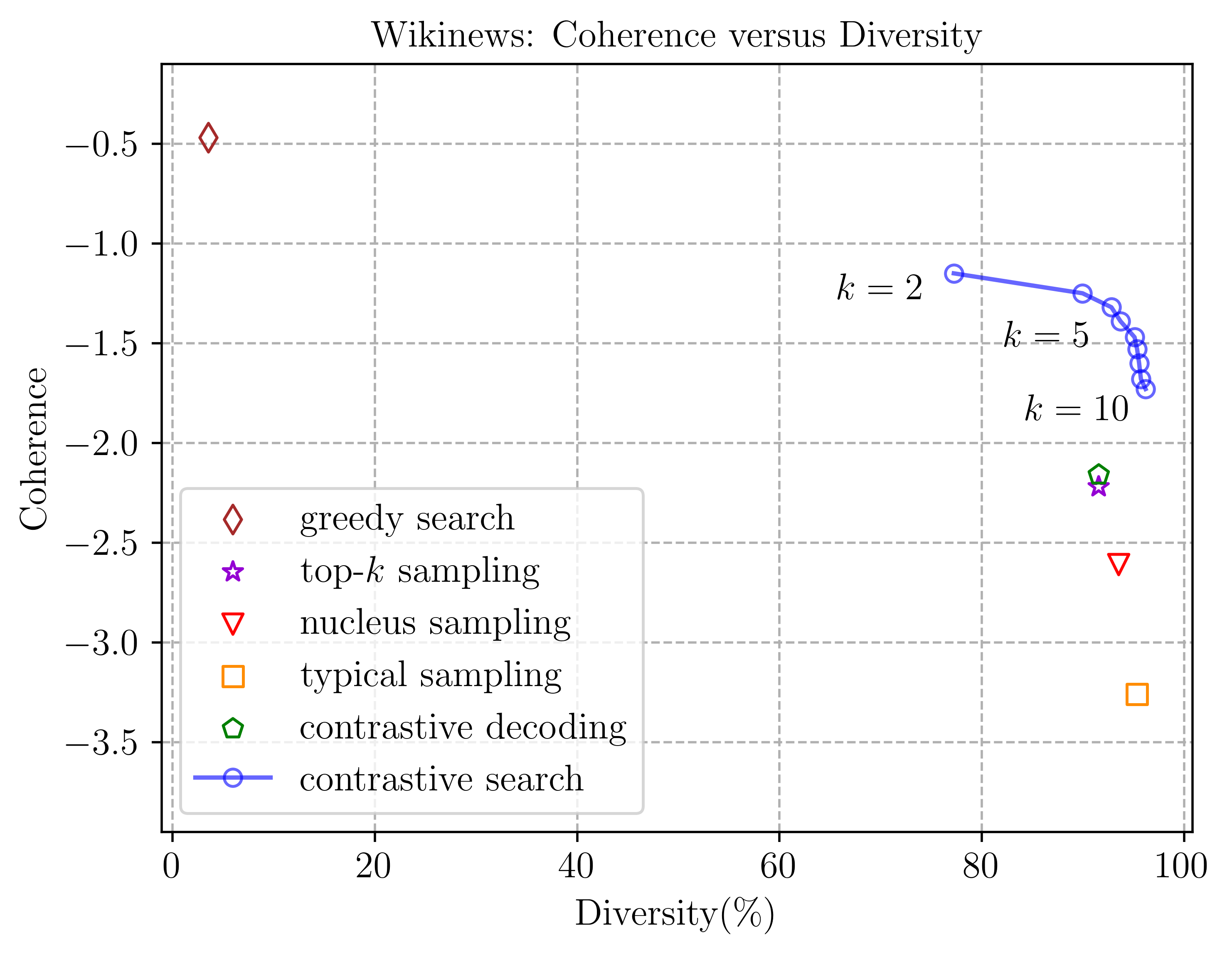}
  \captionof{figure}{\textbf{Wikinews} - Coherence versus Diversity.}
  \label{fig:wikinews_coherence_vs_diversity}
\end{minipage}
\end{figure}

\begin{figure}[h]
\centering
\begin{minipage}{.5\textwidth}
  \centering
  \includegraphics[width=1.\linewidth]{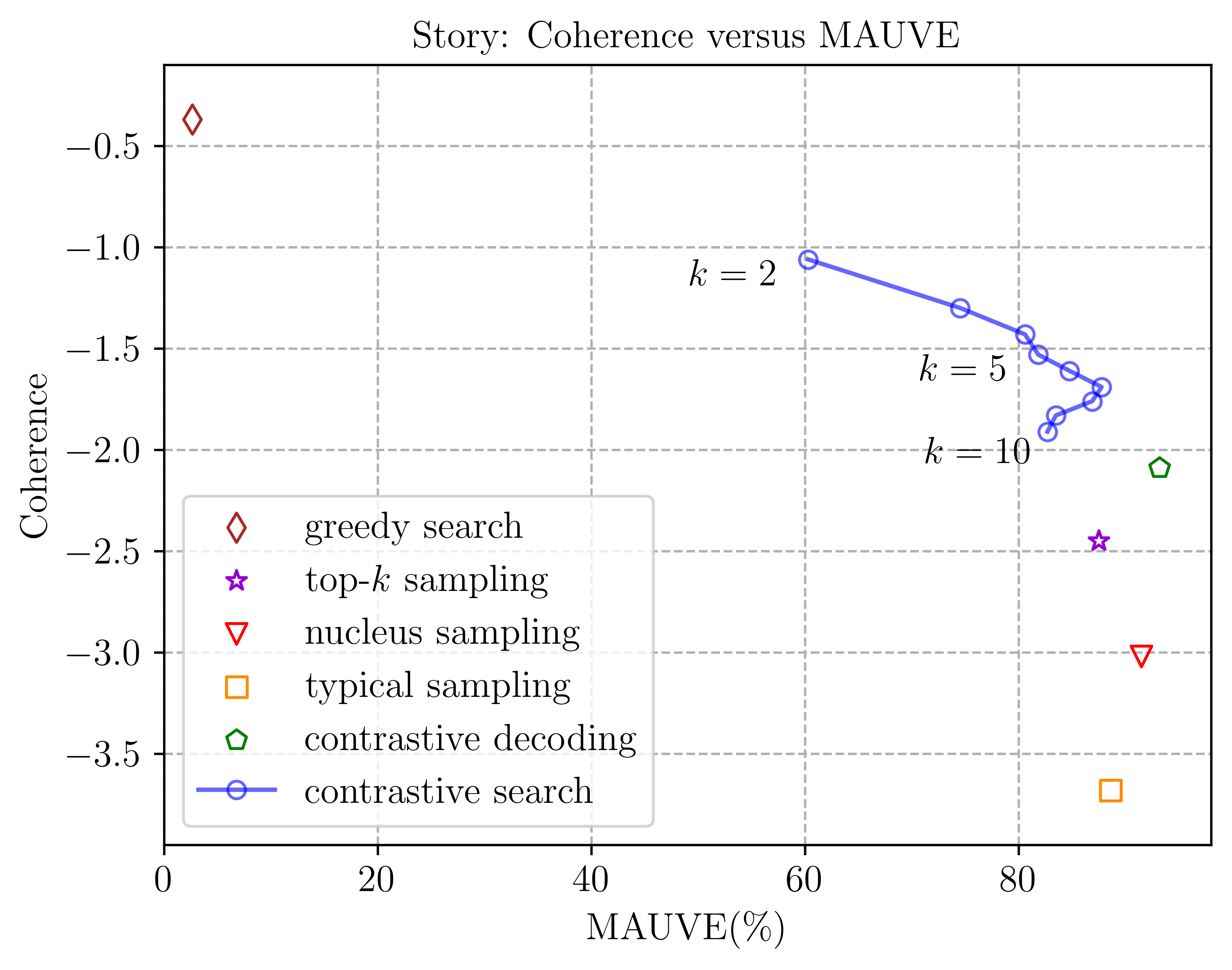}
  \captionof{figure}{\textbf{Story} - Coherence versus MAUVE.}
  \label{fig:story_coherence_vs_mauve}
\end{minipage}%
\begin{minipage}{.5\textwidth}
  \centering
  \includegraphics[width=1.\linewidth]{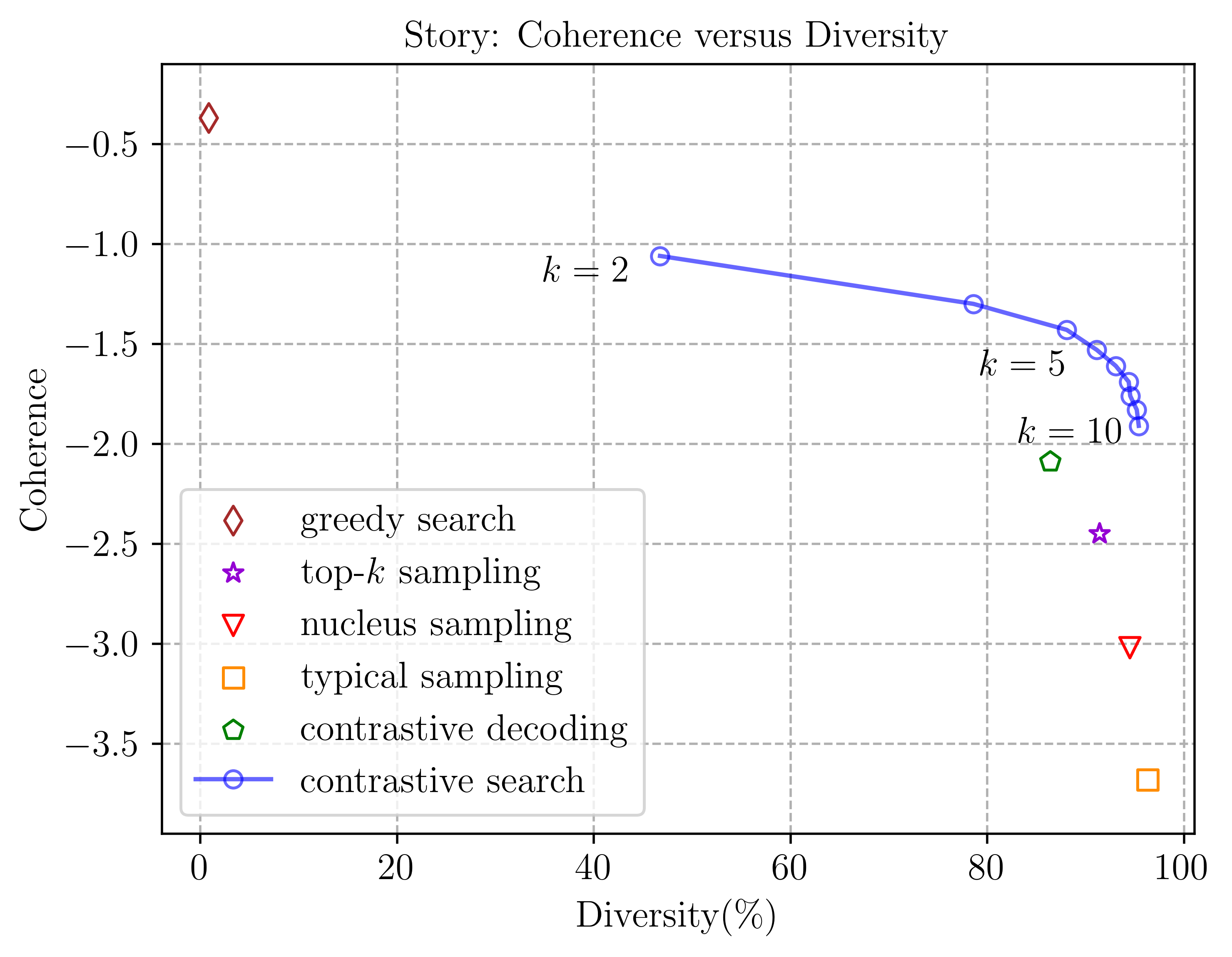}
  \captionof{figure}{\textbf{Story} - Coherence versus Diversity.}
  \label{fig:story_coherence_vs_diversity}
\end{minipage}
\end{figure}

The in-depth analysis results on benchmarks from the news and story domains are presented from Figure~\ref{fig:wikinews_coherence_vs_mauve} to Figure~\ref{fig:story_coherence_vs_diversity}. From the results, we can draw the same conclusion as in~\cref{sec:analysis}.

\end{document}